\documentclass[a4paper, 10pt, twocolumn, twoside]{article}

\usepackage{ISARC}

\usepackage{lscape}
\usepackage{hologo}
\usepackage{acronym}
\usepackage{siunitx}
\usepackage{xurl}
\usepackage{tikz}
\usepackage{graphicx}
\usetikzlibrary{arrows,positioning}

\acrodef{MARA}{Mobile Agile Robotic Assembly}
\acrodef{TCP}{Tool Center Point}
\acrodef{DoF}{Degree of Freedom}
\acrodef{MCR}{mobile construction robot}
\acrodef{ROS}{Robot Operating System}
\acrodef{SLAM}{Simultaneous Localization and Mapping}
\acrodef{CAD}{Computer-Aided Design}
\acrodef{SDK}{Software Development Kit}
\acrodef{IQR}{Interquartile Range}
\acrodef{APE}{Absolute Pose Error}
\acrodef{RMSE}{Root Mean Squared Error}

\newcommand{\norm}[1]{\left\lVert#1\right\rVert}
\newcommand{\vect}[1]{\ensuremath{\mathbf{#1}}}

\begin{document}

\linespread{0.5}

\title{Towards Autonomous Bricklaying Using Boston Dynamics Spot}

\author{Karl-Johan Sorensen$^{1*}$, Nico vom Hofe$^{2*}$ and Frank H.P. Fitzek$^{2,3}$}

\affiliation{
$^1$Department of Architecture, Massachusetts Institute of Technology, United States\\
$^2$Deutsche Telekom Chair of Communication Networks, TU Dresden, Germany\\
$^3$Centre for Tactile Internet with Human-in-the-Loop (CeTI), Dresden, Germany\\
$*$authors contributed equally
}

\email{
  \href{mailto:karljo@mit.edu}{karljo@mit.edu},
  \href{mailto:nico.vom_hofe@tu-dresden.de}{nico.vom\_hofe@tu-dresden.de},
  \href{mailto:frank.fitzek@tu-dresden.de}{frank.fitzek@tu-dresden.de}
}

\maketitle 
\thispagestyle{fancy} 
\pagestyle{fancy}

\begin{abstract}
Mobile legged robots offer unique advantages for on-site construction, including the ability to traverse unstructured terrain, climb stairs, and operate independently between spatially separated work zones. This paper presents a proof-of-concept system for autonomous brick stacking using a Boston Dynamics Spot quadrupedal robot equipped with a manipulator arm. The system operates between a supply area and an assembly area, using AprilTag fiducials for localization and autowalk missions for navigation. Target brick positions and orientations are specified in CAD software and exported directly for robotic execution. We evaluate placement accuracy through ten repeated assemblies of a curved four-layer wall comprising 14 wooden bricks. Results show mean translation errors below 10\,mm and mean yaw errors below 1.5\textdegree{}, while the overall RMSE across all runs is 27.72\,mm (translation) and 2.06\textdegree{} (rotation). The system was also tested outdoors with stairs between the two areas. The workflow is modular and easily extendable to different wall geometries.

\end{abstract}


\section{Introduction}
\label{sec:Introduction}
The construction industry is undergoing a transition towards greater automation, driven by rising housing demand, skilled labor shortages, and persistent safety concerns on construction sites. Yet, compared to sectors such as automotive and logistics, construction has seen limited productivity gains, with annual improvement rates of only \SI{1}{\percent} against an industry average of \SI{2.8}{\percent}~\cite{barbosa_reinventing_2017}. On-site robotic automation has been proposed as one means to address these challenges~\cite{autonomous_construction_robots_melenbrink}.
Despite this potential, the deployment of robotic systems on construction sites remains limited~\cite{zeng_autonomous_2024}. Existing approaches face significant constraints. Large gantry systems require infrastructure that often exceeds the scale of the structure being built. Wheeled and tracked mobile platforms struggle with rough terrain, stairs, and cluttered site conditions. These constraints have confined most successful automation efforts to off-site prefabrication; yet construction fundamentally remains an on-site process where varying conditions resist complete pre-planning.
Recent advances in commercial legged robotics have produced platforms capable of traversing stairs, uneven terrain, and human-centric spaces~\cite{boston_dynamics_spot}. This makes them suited to the multi-level, cluttered environments of real construction sites, where wheeled and tracked platforms face fundamental access constraints. Prior work has demonstrated legged manipulation for general pick-and-place tasks~\cite{kasaragod_dynamic_2024, zhang_learning_2025}, but application to construction assembly has not been explored. To the best of our knowledge, no previous study has quantitatively evaluated the accuracy with which a legged robot can execute a brick-stacking task.

This paper presents a proof-of-concept for what we term the \ac{MARA} workflow: a distributed construction process in which a legged mobile manipulator autonomously traverses between spatially separated supply and assembly areas, picking and placing bricks according to a specified layout. The approach is evaluated using a Boston Dynamics Spot quadrupedal robot equipped with a manipulator arm. Target brick positions are modeled in Rhino, a \ac{CAD} program, and exported directly for robotic assembly. The system localizes by referencing each manipulation operation to a locally visible fiducial marker, thereby decoupling placement accuracy from global odometry drift. Placement accuracy is evaluated through ten repeated assemblies of a curved four-layer wall, reporting translational and rotational errors across layers.

\section{Background}
\label{sec:Background}
The concept of robotic masonry has evolved from early mechanized systems to the dedicated commercial bricklaying systems of today. Early mobile systems such as the BRONCO demonstrated the first robotic bricklaying on a mobile platform but required pre-surveyed sites, remote control, and continuous operator supervision~\cite{pritschow_technological_1996}. Later, feasibility studies formalized the relationship between laying patterns, manipulator trajectories, and achievable precision through planning and optimization~\cite{yu_feasibility_2009}. Although commercial systems such as the Semi-Automated Mason (SAM100) demonstrate viable high-throughput automation, they remain optimized for structured workspaces and require strict workflow integration and supervision by human crews~\cite{madsen_sam100_2019}. These systems also share a spatial constraint: materials must be staged within or immediately adjacent to the robot's workspace. Deploying these systems in real-site contexts with environmental constraints highlights the gap between laboratory robotics and construction realities~\cite{xu_automatic_2019}.
Competition-driven research further validates the autonomy necessary to assemble wall patterns with differently sized and colored bricks within time constraints~\cite{basiri_autonomous_2021,stibinger_mobile_2021,lenz2022nimbro}. Among these, Lenz et al.~\cite{lenz2022nimbro} provide the most detailed quantitative evaluation, reporting end-to-end translational placement errors of \SIrange{13}{19}{\milli\meter} and rotational errors below \SI{1.3}{\degree} using a wheeled omnidirectional platform with a UR10e arm. However, all of these systems depend on wheeled or tracked mobility, restricting their ability to navigate rough terrain, stairs, and cluttered environments. Additionally, despite their mobility, these platforms require close proximity to material supplies or human collaborators for resupply.

Legged robotic platforms can traverse stairs, uneven terrain, and constrained spaces, enabling autonomous operation between spatially decoupled supply and assembly zones without the site layout constraints imposed by material co-location. Early work has explored small-scale humanoid robots for on-site floor tiling, demonstrating perception-driven placement with a NAO biped~\cite{jung_towards_2014}. More recent work on quadruped-mounted manipulators demonstrates pick-and-place pipelines integrating object detection with real-time control~\cite{kasaragod_dynamic_2024}. Learning-based approaches show that multi-stage, long-horizon pick-and-place tasks on legged mobile manipulators can be made robust through sim-to-real transfer~\cite{zhang_learning_2025}. These advances demonstrate the growing capability of legged manipulation; however, application to masonry assembly remains unexplored.

This paper addresses that gap, combining two domains that, to the best of our knowledge, have not previously been merged in this context: legged manipulation~\cite{kasaragod_dynamic_2024, zhang_learning_2025} and mobile construction robotics~\cite{zeng_autonomous_2024, helm_mobile_2012, dorfler_mobile_2016, giftthaler_mobile_2017, stibinger_mobile_2021}. We present the first demonstration of autonomous brick stacking on a legged mobile manipulator and provide the first quantitative evaluation of placement accuracy for a legged robot performing construction assembly. Our proposed system introduces a \ac{CAD}-to-execution pipeline integrated with a legged manipulator that can operate across spatially decoupled supply and assembly areas connected by stairways, which are inaccessible to wheeled or tracked platforms.

\section{Methodology}
\label{sec:methodology}
\subsection{System overview}

\begin{figure}[htb]
    \centering
    \begin{subfigure}[t]{.49\textwidth}
        \centering
        \includegraphics[width=\textwidth]{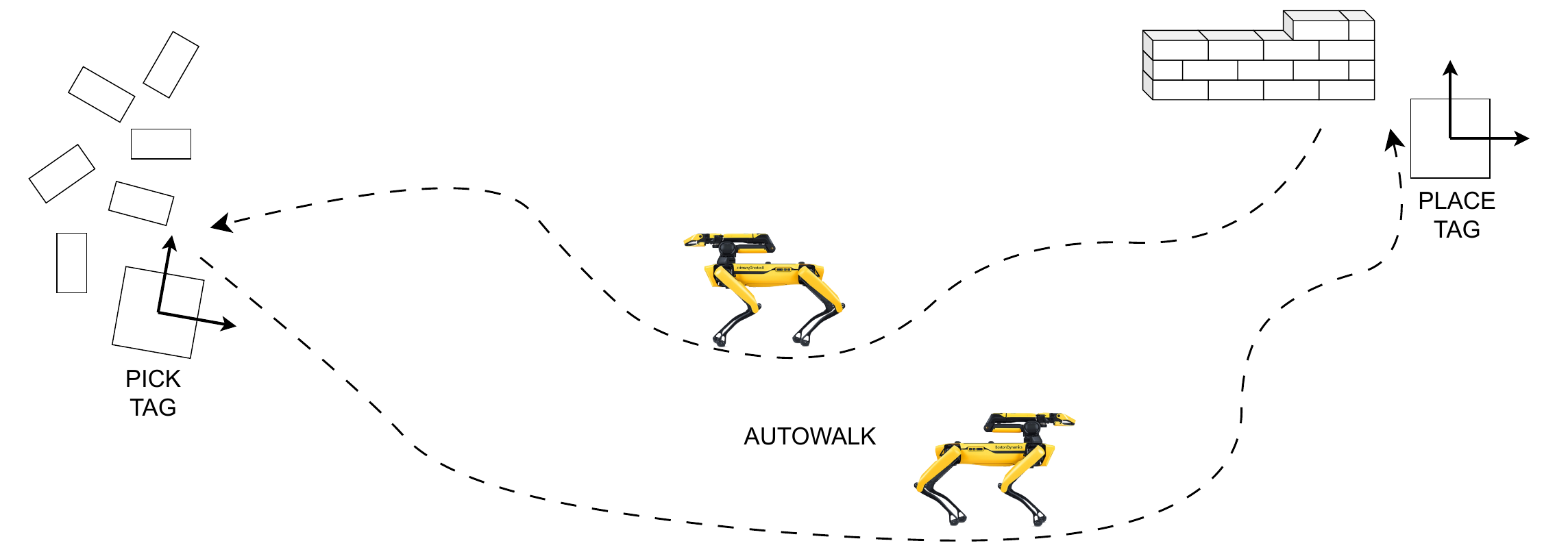}
        \caption{Pick-place cycle between supply and assembly areas}
        \label{fig:pick-place-overview}
    \end{subfigure}
    \par\medskip
    \begin{subfigure}[t]{.42\textwidth}
        \centering
        \resizebox{\textwidth}{!}{%
            \begin{tikzpicture}[
inner/.style={draw, inner sep=4},
outer/.style={draw,inner sep=4, outer sep=2}
]

\node[outer, minimum height=2cm, minimum width=7.5cm] (control) {};
\node[rectangle, outer sep=3, anchor=north] (controllabel) at (control.north) {Main Sequence Controller};

\node[inner, text height=3ex, anchor=north west, xshift=0.4cm, yshift=-0.7cm, align=center] (pick) at (control.north west) {Pick\\Module};
\node[inner, text height=3ex, anchor=west, align=center] (autowalktarget) at ([xshift=0.25cm]pick.east) {Autowalk\\to Target};
\node[inner, text height=3ex, anchor=west, align=center] (place) at ([xshift=0.25cm]autowalktarget.east) {Place\\Module};
\node[inner, text height=3ex, anchor=west, align=center] (autowalksource) at ([xshift=0.25cm]place.east) {Autowalk\\to Source};

\def\xoffset{2cm}
\def\yoffset{2cm}
\def\arrowdist{3mm}

\node[outer, minimum height=1.2cm, minimum width=2cm] 
    (cad) at ([xshift=-\xoffset, yshift=\yoffset]control.north) {};

\node[rectangle, outer sep=2, anchor=north] (cadlabel) at (cad.north) {CAD};
\node[rectangle,draw,minimum width=0.6cm,minimum height=0.2cm, anchor=north] (b1) at (cadlabel.south) {};
\node[rectangle,draw,minimum width=0.6cm,minimum height=0.2cm, anchor=north] (b2l) at (b1.south west) {};
\node[rectangle,draw,minimum width=0.6cm,minimum height=0.2cm, anchor=north] (b2r) at (b1.south east) {};

\node[outer, minimum height=1.2cm, minimum width=2cm] 
    (dt) at ([xshift=\xoffset, yshift=\yoffset]control.north) {};

\node[rectangle, outer sep=2, anchor=north] (dtlabel) at (dt.north) {Digital Twin};

\coordinate (control_north_left)  at ([xshift=-\xoffset]control.north);
\coordinate (control_north_right) at ([xshift=\xoffset]control.north);

\draw[->] (cad.south) -- (control_north_left) 
    node[midway, left, align=center] {source and target\\brick poses};

\draw[->] ([xshift=\arrowdist]control_north_right) -- ([xshift=\arrowdist]dt.south);
\draw[->] ([xshift=-\arrowdist]dt.south) -- ([xshift=-\arrowdist]control_north_right);

\draw[->] (pick) -- (autowalktarget);
\draw[->] (autowalktarget) -- (place);
\draw[->] (place) -- (autowalksource);
\draw[->] (autowalksource.south) -- ++(0,-0.25) -| (pick.south);

\end{tikzpicture}
        }
        \caption{Software pipeline from CAD to robot execution}
        \label{fig:system-overview}
    \end{subfigure}
    \caption{System overview}
    \label{fig:system-and-pick-place-overview}
\end{figure}

Our autonomous on-site construction system utilizes a \textit{Boston Dynamics} Spot robot with a six-\ac{DoF} manipulator arm, offering a reach of \SI{984}{\milli\meter} at full extension and a lifting capacity of \SI{11}{\kilo\gram}~\cite{spotarm}. This robot was chosen for its off-the-shelf availability, which makes deployment significantly more feasible than developing and certifying a custom platform. The system comprises two distinct areas: a supply area containing bricks and an assembly area where the target structure is constructed. Each area is defined by an AprilTag~\cite{apriltag} fiducial from the \textit{Tag36h11} set, as illustrated in Figure~\ref{fig:pick-place-overview}. Specifically, the \textit{PICK-TAG} is situated in the supply area, while the \textit{PLACE-TAG} is located within the assembly area. Unlike systems that depend on active positioning infrastructure such as total stations, UWB, or long-range RFID, which require external power and increase installation effort, the localization system relies entirely on passive AprilTags that need neither a power supply nor geometric pre-calibration.

All brick poses are specified relative to these fiducial frames, providing robustness against cumulative drift in odometry. In the context of commanding arm motions, poses are transformed from the fiducial frame to Spot's vision frame. The latter is a world-fixed frame maintained by the robot's visual localization system~\cite{spotframes}. By referencing each operation to the fiducial, this approach decouples placement accuracy from global localization drift.

The target brick positions are 3D modeled in the \ac{CAD} software, Rhino. A custom script was developed to extract position and orientation data. The script then exports JSON configuration files that specify the poses of the bricks relative to the fiducials. This process establishes a direct connection between architectural design and robotic execution. Additionally, a human operator recorded two autowalk missions for the supply-to-assembly and return journeys using \textit{Boston Dynamics'} autonomous navigation system.

The process of repeating sequences for each brick configuration is illustrated in the software pipeline in Figure~\ref{fig:system-overview}. During execution, the system iterates through the configuration file and executes a four-step cycle for each brick: (1) pick a brick from the supply area, (2) walk to the assembly area, (3) place the brick at the target pose, and (4) return to the supply area. The digital twin component enables a remote operator to monitor the robot's state and perception in real time (Section~\ref{sec:digital_twin}).

The system was developed using Python and the \textit{Boston Dynamics} \ac{SDK}. The source code and experimental data are publicly available at \url{https://github.com/soerensenkarl/MARA-bricklaying}.

\begin{figure}[htb]
    \centering
    \includegraphics[width=0.75\linewidth,trim=0cm 0cm 0cm 0cm,clip]{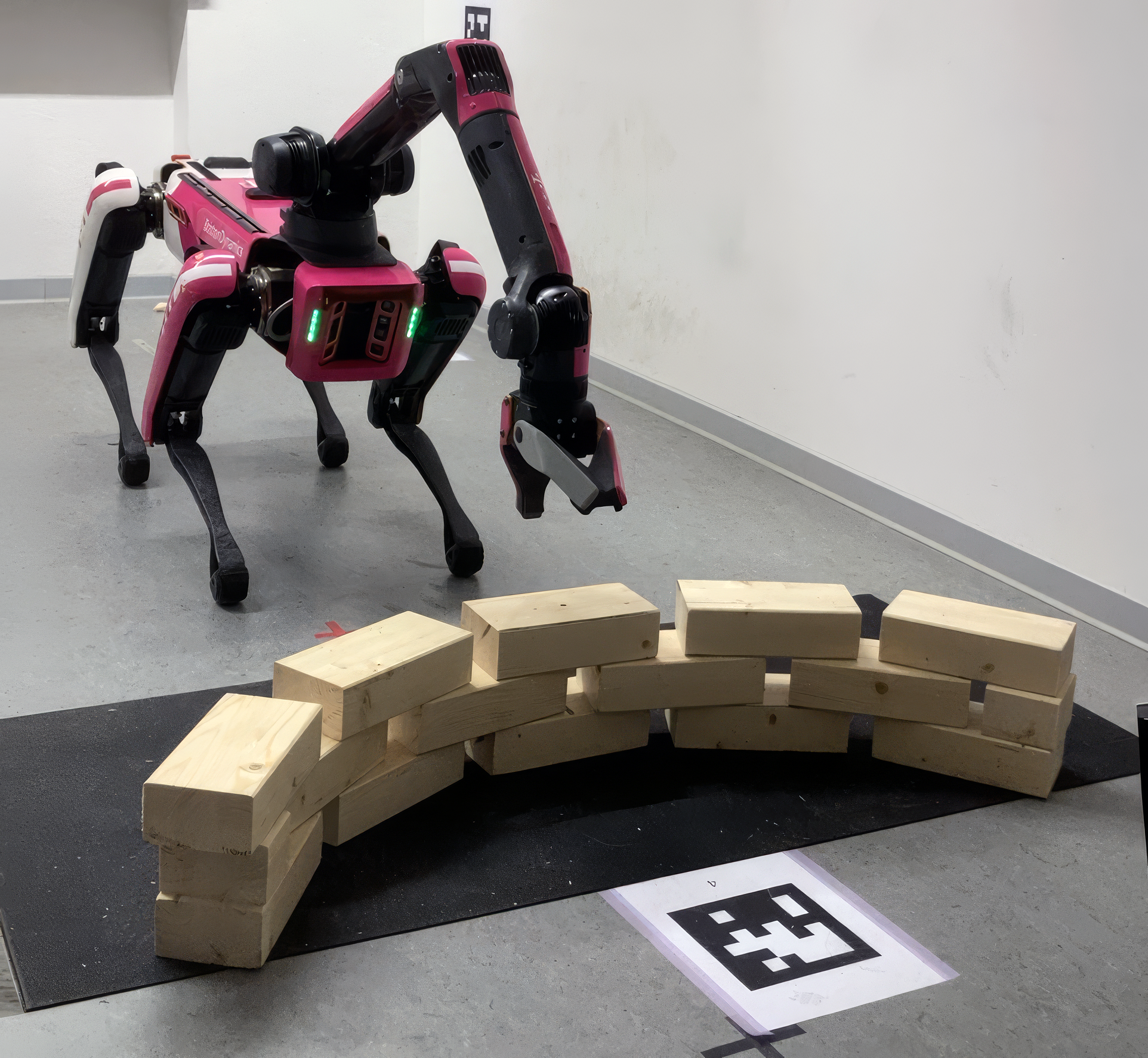}
    \caption{Spot after placing a brick on a curved wall}
    \label{fig:placing_brick}
\end{figure}

\subsection{Manipulation}
A common positioning strategy is employed in both pick and place operations. The robot first detects the relevant fiducial and moves its base to a position \SI{0.8}{\meter} behind the target brick along the fiducial's x-axis, aligning its heading with the positive x-direction of the fiducial. This pre-positioning ensures the arm has sufficient reach to access the brick and adequate visibility of the fiducial markers. After reaching this position, the system reacquires the fiducial transform to establish a stable reference for arm commands.

The pick sequence proceeds as follows: The gripper opens and the arm approaches the brick, moving the hand frame to \SI{20}{\centi\meter} above the target position specified in the configuration file. It then descends to \SI{3.5}{\centi\meter} above the target, pauses briefly, closes the gripper to secure the brick, and lifts to a transport height of \SI{80}{\centi\meter}. The gripper maintains a pitch angle of \SI{95}{\degree} (nearly vertical, pointing slightly backward) during picking. This angle was empirically determined to provide a reliable grasp.

The place sequence inverts this operation. The arm approaches from \SI{30}{\centi\meter} above the target position, descends to \SI{4.5}{\centi\meter} above the surface, pauses, opens the gripper to release the brick, retracts, and stows. During placement, the end-effector pitch is set to \SI{90}{\degree} (vertical) and the yaw is set according to the target orientation specified in the configuration file, enabling non-axis-aligned placements for curved geometries. A z-offset of \SI{-1.5}{\centi\meter} compensates for the distance from the hand frame to the center of the gripped brick. Hip height assist is enabled throughout both sequences to permit base height adjustments.

\subsection{Navigation}

The robot autonomously navigates between pick and place operations, executing pre-recorded autowalk missions to traverse the supply and assembly areas. Each autowalk consists of a graph of waypoints and edges, along with visual snapshots that were captured during the recording process. During operation, the module uploads the graph to Spot's GraphNav system and begins localization by searching for the nearest visible AprilTag. During navigation, GraphNav continuously localizes the robot by combining odometry with visual and geometric features and by comparing real-time sensor data against stored map snapshots~\cite{graphnav}. The robot then autonomously navigates along the recorded path while avoiding obstacles.

\subsection{Remote supervision}
\label{sec:digital_twin}

A digital twin component built on \ac{ROS}~2 Humble enables real-time remote supervision of the robot during operation. The wrapper streams the robot's internal state and visual perception to a remote operator via RViz. The streamed state includes the battery level and the position of each joint, allowing the robot's full pose to be reconstructed at each timestamp. The robot's visual perception of its environment is rendered as six point cloud streams from the five body-mounted RGB-D cameras and the gripper camera. This provides an operator with an intuitive, egocentric view of the robot's surroundings, as opposed to raw camera feeds. Environmental observations can be gradually integrated into a single 3D representation of the site using the mapping framework described in~\cite{spot_mapping}. The digital twin achieves a mean frame rate of \SI{3.5}{\hertz} for the RGB-D streams and approximately \SI{50}{\hertz} for joint state updates. Each camera transmits a time-synchronized pair of RGB and depth \ac{ROS} messages, with mean sizes of \SI{921.7}{\kilo\byte} and \SI{614.46}{\kilo\byte}, respectively.

\subsection{Experimental setup}

\begin{figure}[htb]
    \centering
    \includegraphics[width=\linewidth,trim=0cm 0cm 0cm 10cm,clip]{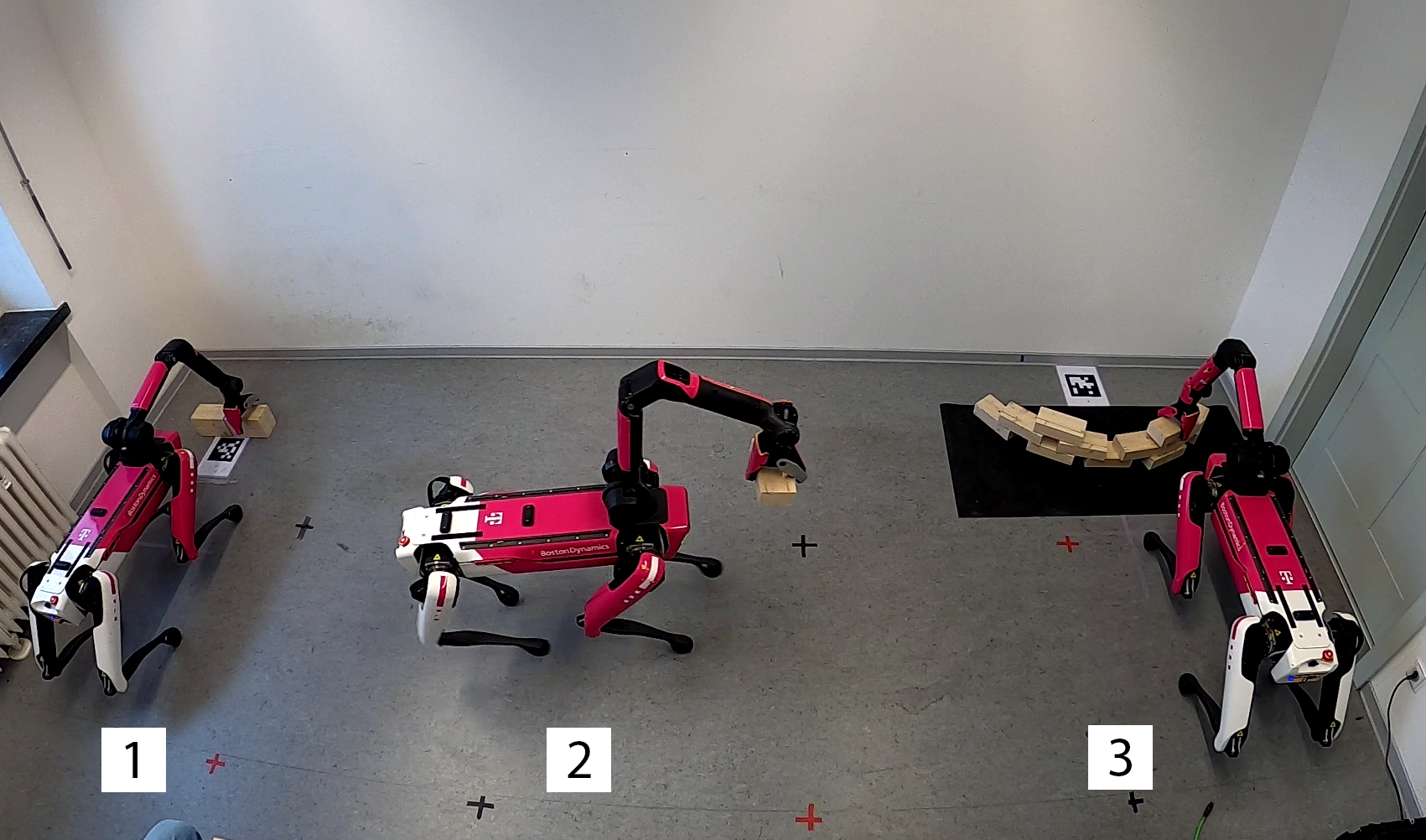}
    \caption{Superimposed video frames from the experiment showing Spot at pick position (1), autowalking (2), and at place position (3)}
    \label{fig:experimental_setup}
\end{figure}

The experimental site comprised a supply area and an assembly area, separated by a distance of approximately \SI{4}{\meter}. Fiducials were placed in opposite corners of the room. Figure \ref{fig:experimental_setup} shows the experimental room-scale setup, including the three stages of picking up, walking with, and placing bricks. To avoid potential damage to the robot, wooden blocks corresponding to the standard brick proportions (\SI{200}{\milli\meter} \(\times\) \SI{100}{\milli\meter} \(\times\) \SI{60}{\milli\meter}) were used instead of actual bricks. These wooden bricks have a mean weight of \SI{550}{\gram}.

In the supply area, the bricks were arranged in two vertical stacks of seven bricks each.  The target structure was a curved wall with four layers and a total of 14 bricks. 

There were five bricks in the base layer and fewer in each subsequent layer as the wall tapered toward the top. A non-slip mat was positioned beneath the base layer to mitigate the risk of slippage during the process of placing the upper layers.

\section{Results}
\label{sec:Results}
The evaluation of placement accuracy was conducted by selecting planar position error (x and y) and yaw error as the primary metrics. Errors in the remaining pose dimensions (z, roll, and pitch) would indicate an unsuccessful build and were not analyzed. Any structural collapse resulting from substantial placement inaccuracies was designated as a failure. All ten experimental runs completed successfully without failure.

\begin{figure}[htb]
    \centering
    \includegraphics[width=.8\linewidth]{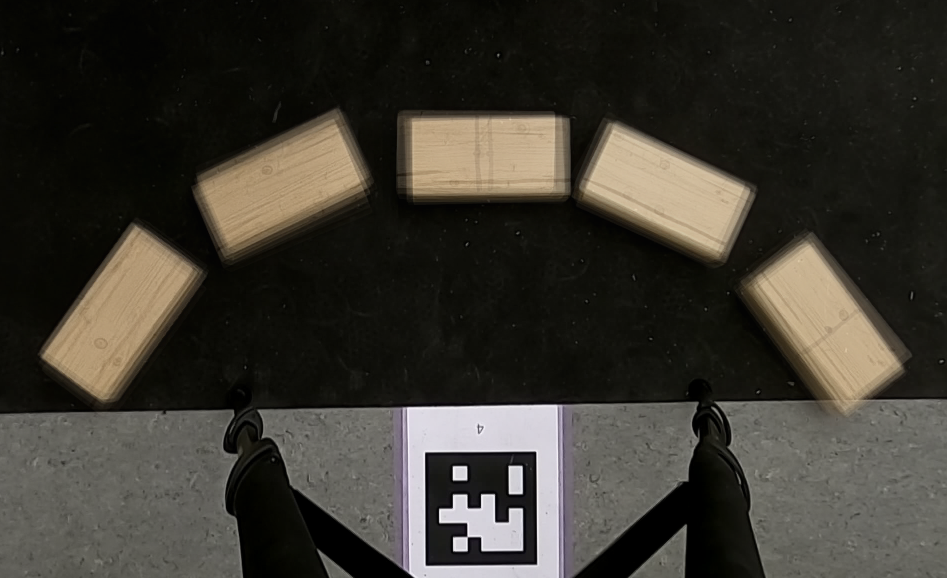}
    \caption{Superimposed overlay of all ten first-layer placements, showing variability across runs}
    \label{fig:overlay}
\end{figure}

To record brick positions, a downward-facing camera was mounted at a height of approximately \SI{1.33}{\meter} above the \textit{PLACE-TAG}. Reproducing the camera position and parameters in a \ac{CAD} model enabled the generation of ground-truth images for each layer of the target wall. A layer-wise image overlay between recorded images and ground truth was used to calculate the planar center-point offset between each placed brick and its target position. The same overlay was used to extract brick contours, enabling measurement of the angular deviation between desired and actual orientations, recorded as yaw error.
\begin{figure}[tb]
    \centering
    \begin{subfigure}[t]{.48\textwidth}
        \centering
        \includegraphics[clip, trim={0.7cm 1.5cm 0.5cm 2cm}, width=\linewidth]{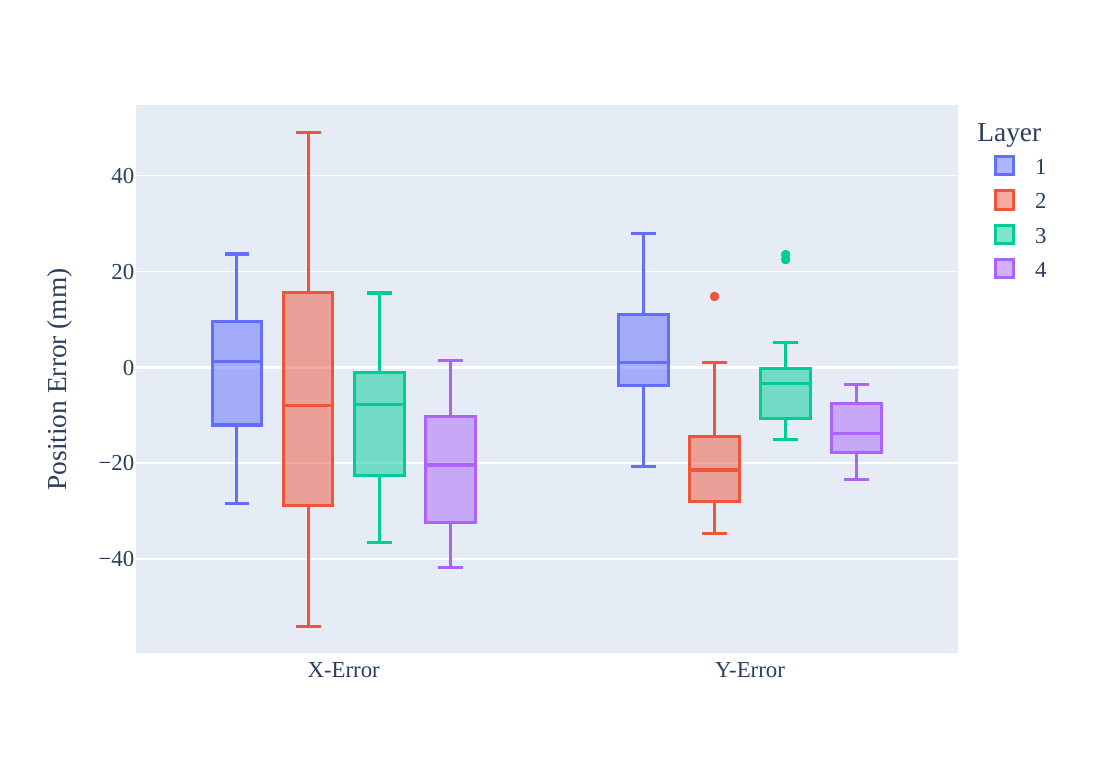}
        \caption{Position error for each layer in x and y direction}
        \label{fig:xyerr}
    \end{subfigure}
    \hfill
    \begin{subfigure}[t]{.48\textwidth}
        \centering
        \includegraphics[clip, trim={1cm 1.5cm 0.5cm 1cm}, width=\linewidth]{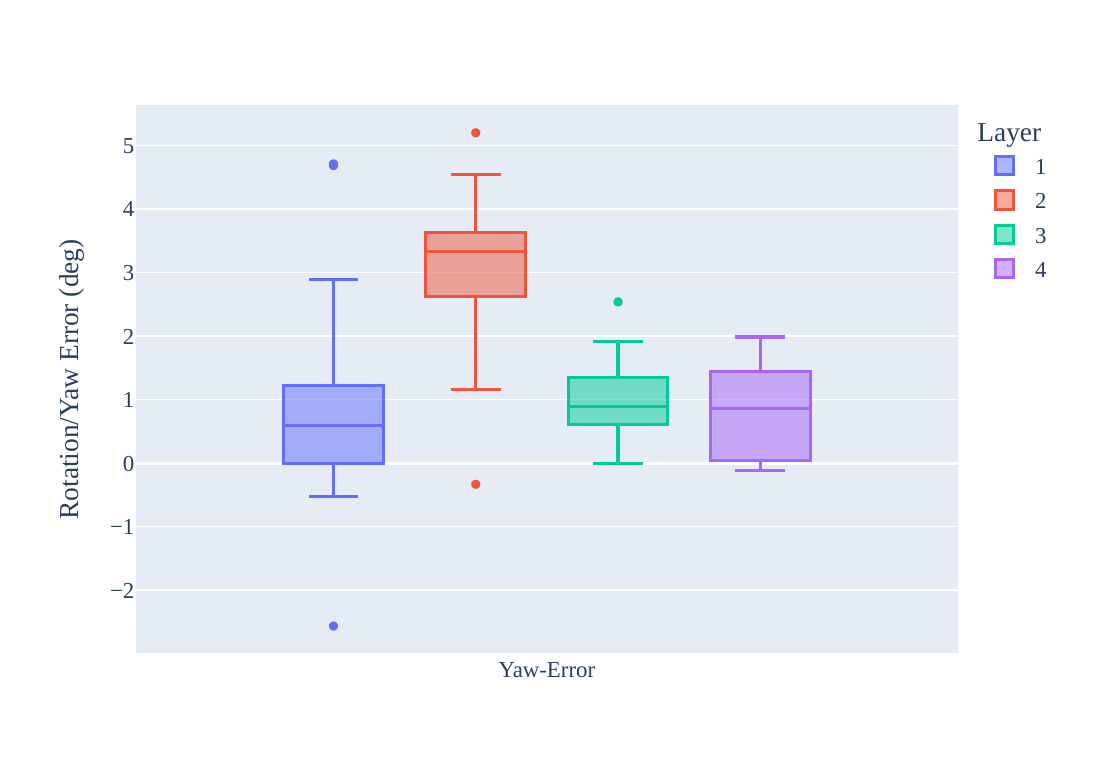}
        \caption{Yaw error for each layer}
        \label{fig:yawerr}
    \end{subfigure}
    \caption{Error metrics for the individual layers}
    \label{fig:layererrors}
\end{figure}

The evaluation comprised ten fully autonomous constructions of the same structure. Figure~\ref{fig:layererrors} presents the resulting pose errors, grouped by construction layer. Error distributions are visualized using boxplots based on quartiles ($Q_1$, $Q_2$, $Q_3$) and the \ac{IQR}, defined as $Q_3 - Q_1$. The whiskers extend to the most extreme data points within $1.5\times\mathrm{IQR}$ from the first and third quartiles. The single points represent maximum values that exceed the upper fence and constitute actual data values that surpass the established threshold. Consequently, these points are categorized as outliers.

The initial layer, comprising five bricks, exhibits the highest number of measurements and the widest range of target yaw angles (\SI{107.8}{\degree}). This range is necessary to establish the curved base of the wall. As shown in Figure~\ref{fig:xyerr}, this layer achieves the smallest median planar position errors, measuring \SI{1.18}{\milli\meter} in the x direction and \SI{1.01}{\milli\meter} in the y direction. Despite the large variation in the target yaw angles, Figure~\ref{fig:yawerr} shows a low median yaw error of \SI{0.6}{\degree} for the initial layer.

The second layer demonstrates a higher prevalence of pose errors compared to the base layer, showing median position offsets in the negative x and y directions and the most significant median yaw error among all layers, measuring at \SI{3.32}{\degree}. The third and fourth layers also exhibit slight negative planar offsets; however, their median yaw errors remain below \SI{0.9}{\degree}.
\begin{equation}
    \mathrm{APE}_i = \norm{\widehat{\vect{p}}_i - \vect{p}_i}
    \label{eq:ape}
\end{equation}
Equation~\ref{eq:ape} defines the metric used to evaluate the \ac{APE} of each brick. Aggregating all brick poses across the ten runs, the resulting \ac{RMSE} is \SI{27.72}{\milli\meter} for the translational component and \SI{2.06}{\degree} for the rotational component.

Several factors contribute to the observed error characteristics. The low error in the first layer can be attributed to the anti-slip mat, which increases friction at the ground level. Friction is much lower in higher layers due to the smooth surfaces of the wooden bricks. Furthermore, Spot is designed for field mobility rather than factory-grade motion repeatability, contributing to inherent positioning variability. The apparent increase in errors at higher layers may also result from greater distance between the gripper and the visual fiducial anchor point, as the fiducial is fixed at ground level while the placement height increases with each layer.

Execution times were recorded by logging timestamps at the start and end of each operation during the assembly sequence. Table~\ref{tab:cycle_times} summarizes the mean durations for each phase of the pick-place cycle.

\begin{table}[!htb]
\centering
\caption{Mean cycle times for each operation phase}
\label{tab:cycle_times}
\begin{tabular}{lr}
\hline
Operation & Duration (s) \\
\hline
Pick & 9.18 \\
Walk to assembly area & 15.46 \\
Place & 9.68 \\
Walk to supply area & 16.15 \\
Total cycle & 50.47 \\
\hline
\end{tabular}
\end{table}

The mean cycle time of \SI{50.5}{\second} per brick results in a total assembly time of approximately 12 minutes for the 14-brick structure with a supply-to-assembly distance of \SI{4}{\meter}. Navigation accounts for 63\% of the total cycle time. These times reflect conservative speed parameters set during development; timeouts, motion durations, and arm trajectories were not optimized for speed, leaving considerable potential for efficiency improvements.

\begin{figure}[!htb]
    \centering
    \begin{subfigure}[t]{0.4\textwidth}
        \centering
        \includegraphics[width=1.0\linewidth]{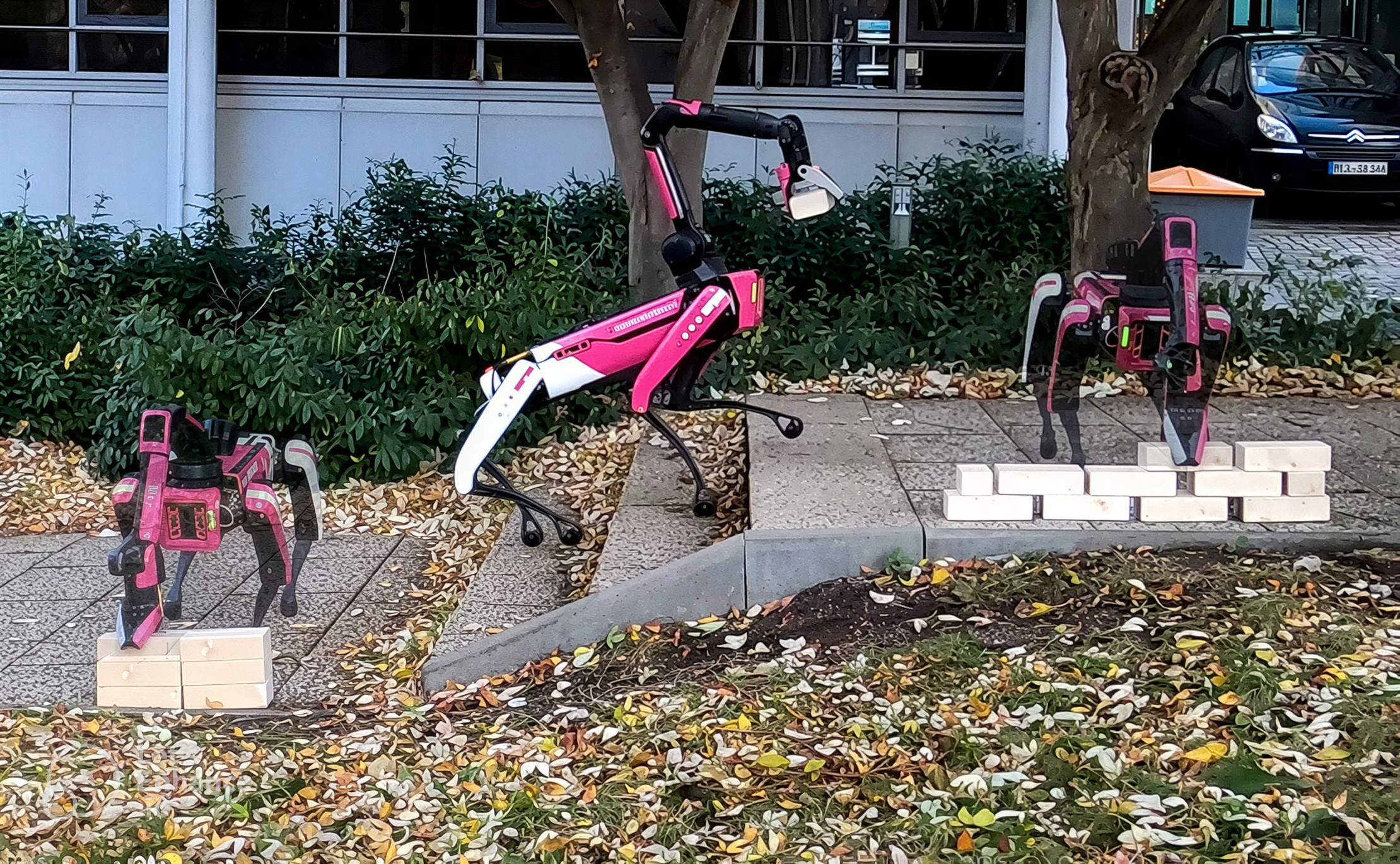}
        \caption{Superimposed frames from outdoor brick-stacking video}
        \label{fig:outdoor_stages}
    \end{subfigure}
    \par\medskip
    \begin{subfigure}[t]{.4\textwidth}
        \centering
        \includegraphics[width=0.8\linewidth]{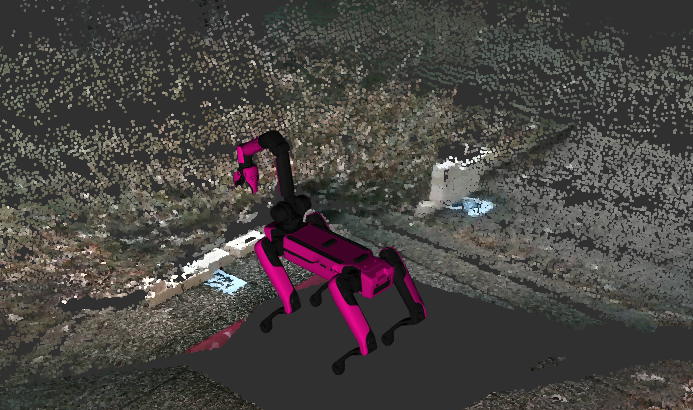}
    \end{subfigure}
    \par\medskip
    \begin{subfigure}[t]{.4\textwidth}
        \centering
        \includegraphics[width=0.8\linewidth]{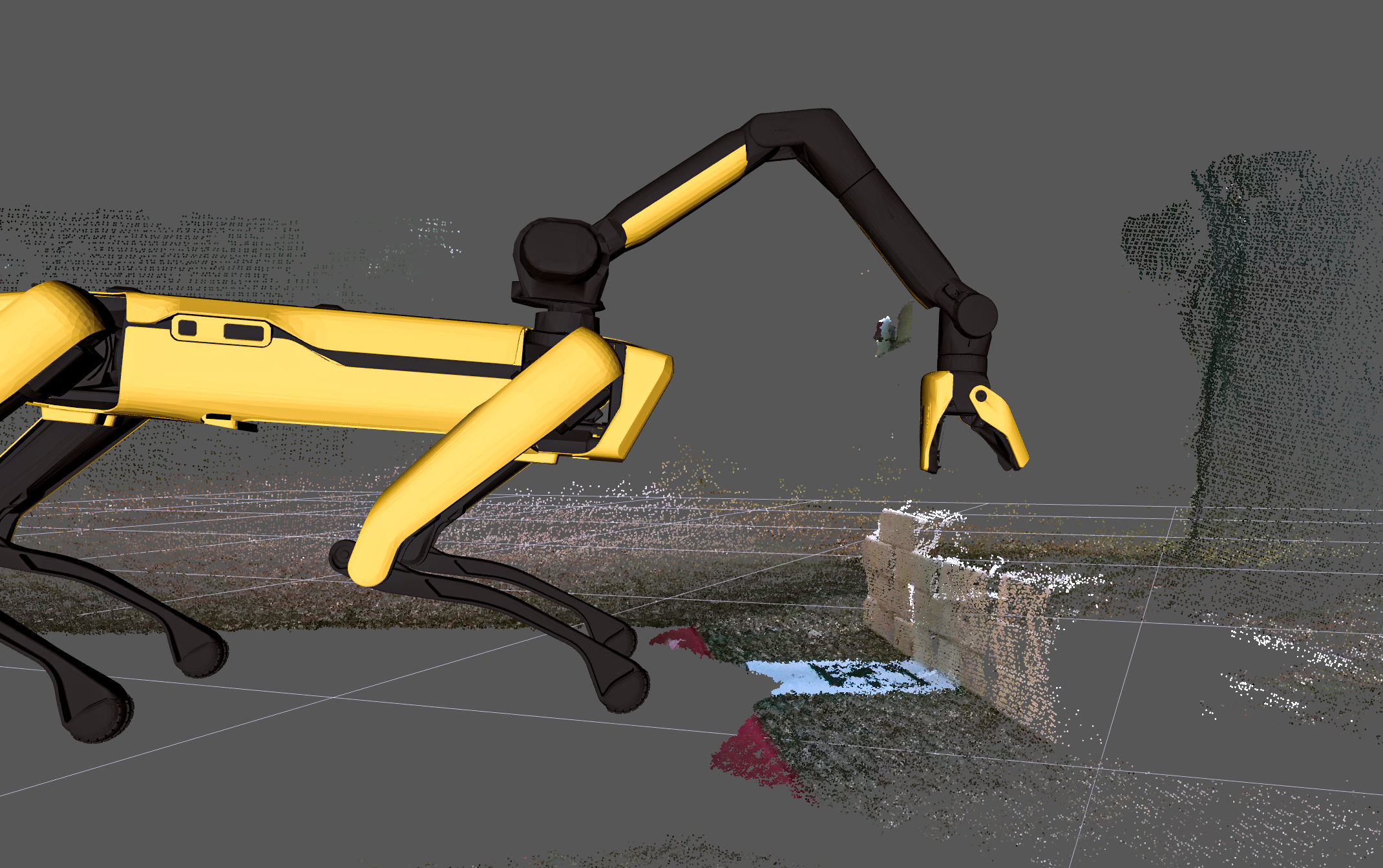}
        \caption{Digital Twin - Real-Time supervision using the ROS2 Wrapper and visualization in RViz}
        \label{fig:outdoor_rviz}
    \end{subfigure}
    \caption{Autonomous construction scenario in an outdoor setting including stairways}
    \label{fig:outdoor_test}
\end{figure}

To contextualize these results, the most directly comparable system is the NimbRo UGV~\cite{lenz2022nimbro}, a wheeled platform with a UR10e arm that reported mean translational placement errors of \SIrange{13}{19}{\milli\meter} and rotational errors of \SIrange{0.45}{1.26}{\degree} in autonomous brick-placement trials. Our mean translational errors below \SI{10}{\milli\meter} are comparable to these figures, while the overall \ac{RMSE} of \SI{27.72}{\milli\meter} reflects sensitivity to occasional outliers at higher layers. The \textit{In situ Fabricator}~\cite{dorfler_mobile_2016, giftthaler_mobile_2017}, a \SI{1.4}{\tonne} tracked platform with an industrial arm offering \SI{2.55}{\meter} reach and \SI{40}{\kilogram} payload, achieved a maximum assembly error of \SI{7}{\milli\meter} over 1600 bricks with substantially different hardware. Other mobile manipulators report grasping accuracies of $\pm$\SIrange{30}{50}{\milli\meter}~\cite{stibinger_mobile_2021,basiri_autonomous_2021} but do not quantify final placement error. No prior work reports placement accuracy for a legged robot in a construction assembly context; the ten repeated assemblies therefore serve as both an absolute accuracy evaluation and an internal repeatability baseline. The manipulation-only cycle time (pick and place, $\approx$\SI{18.9}{\second}) is comparable to other mobile manipulators, while the total cycle time of \SI{50.5}{\second} is dominated by navigation between spatially decoupled work zones, a phase absent in systems where materials are staged within the robot's workspace.

The indoor experiments evaluate placement accuracy under controlled conditions, but the primary motivation for a legged platform is access to terrain that wheeled and tracked systems cannot traverse. To validate this, a second experimental setup was created outdoors with the supply and assembly areas at different elevations, connected by stairways with climbing steps, as shown in Figure~\ref{fig:outdoor_stages}. This configuration demonstrates operation in an environment where wheeled robots would fail. The robot autonomously descended stairs to pick bricks and climbed back to the assembly area to place them. Although individual pose errors were not measured for this setup, repeated runs successfully stacked a planar wall. A continuous video of the outdoor assembly is available at \url{https://youtu.be/p-ExWFgYa-I}.

For remote supervision, the robot's state and perception were continuously streamed to a remote operator via the \ac{ROS}~2 digital twin (Section~\ref{sec:digital_twin}), as shown in Figure~\ref{fig:outdoor_rviz}.

\section{Conclusion and future work}
\label{sec:conclusion}
This paper demonstrates a proof-of-concept for autonomous bricklaying using a legged mobile robot. We introduce a \ac{MARA} workflow deploying a mobile quadrupedal robot with an integrated manipulator arm. This allows the robot to navigate autonomously between a supply area and an assembly area while picking and placing bricks. Unlike stationary or wheeled industrial manipulators, the mobile robot can traverse unstructured environments and climb stairs, making it well-suited for on-site construction.

During our evaluation of the robot's accuracy, it autonomously constructed a curved four-layer wall. It successfully completed all ten experimental runs without human intervention. The system achieved mean translation errors below \SI{10}{\milli\meter} and mean yaw errors below \SI{1.5}{\degree}, demonstrating the viability of legged mobile manipulation for construction tasks.

While the current system focuses on a well-defined operational setup, several aspects provide opportunities for further extension. The use of a single AprilTag fiducial per work area enables reliable localization within the tested scale. Extending to larger structures would require a multi-fiducial configuration. A quantitative evaluation was performed in a controlled indoor environment to ensure reproducibility. This evaluation was complemented by outdoor experiments that demonstrated qualitative feasibility. The current implementation uses dry stacking to provide a simplified baseline for the assembly of the structure. The system design does not preclude the integration of mortar or other bonding materials. With the arm installed, the robot retains a remaining body payload capacity of approximately \SI{6}{\kilogram} \cite{spotarm}, which may accommodate a reservoir for bonding agents, though this has not been tested. Validating this extension with clay bricks and mortar application is a priority for future work.

The system uses a general-purpose, off-the-shelf robotic platform, which offers advantages in terms of availability, robustness, and flexibility. While task-specific design modifications could improve efficiency and practicality in construction scenarios, this level of specialization would increase development and production costs, particularly given the limited use of construction robots. Currently, navigation between supply and assembly areas relies on pre-recorded autowalk missions. Replacing these with fully autonomous path planning remains an open challenge. Additionally, developing recovery strategies for failed grasps or misplacements is a future area of development. 

\begin{figure}[htb]
    \centering
    \includegraphics[width=.6\linewidth,trim=0cm 4cm 0cm 12cm,clip=true]{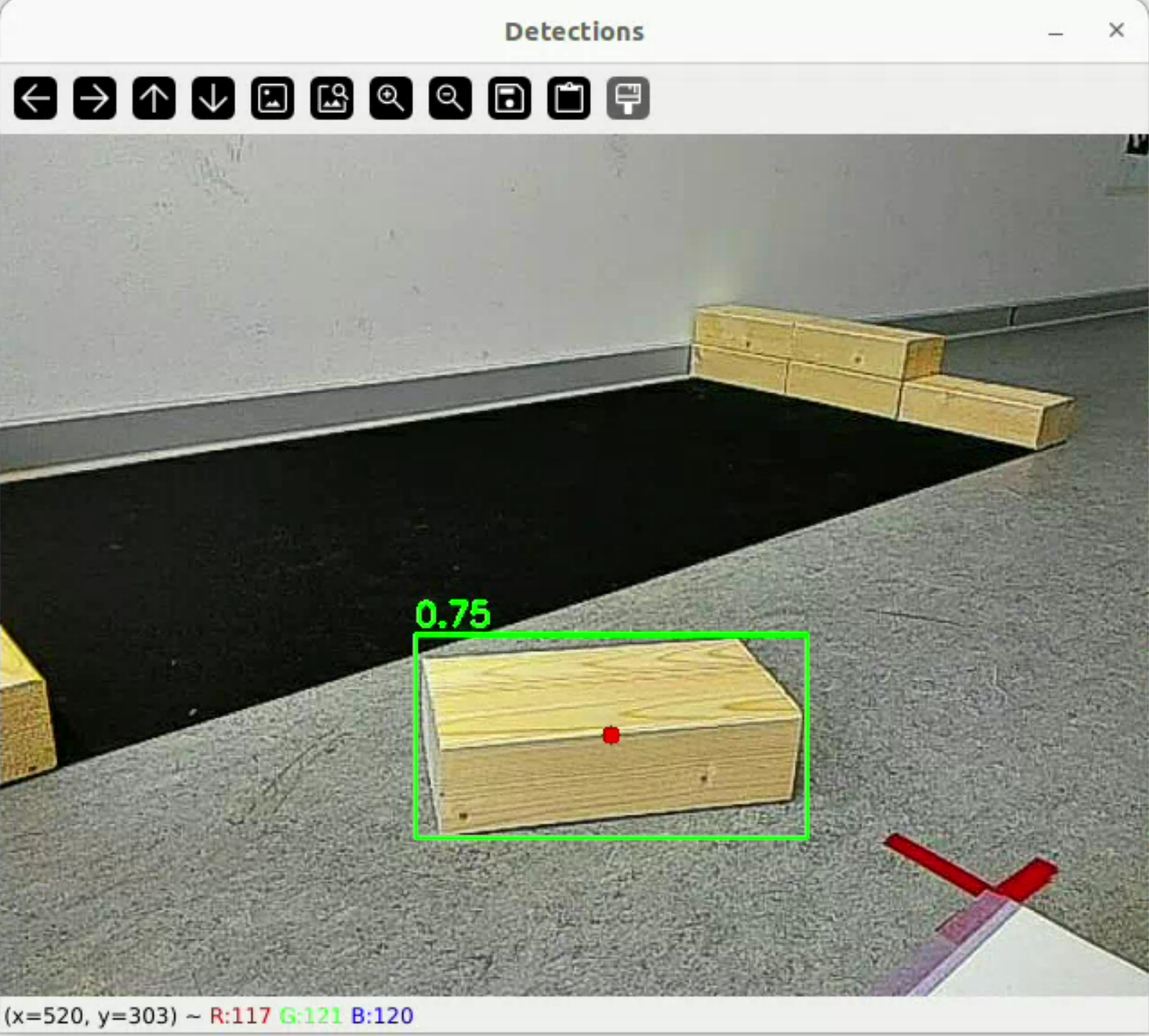}
    \caption{Brick detection model detecting a brick with \SI{75}{\percent} confidence; showing the bounding box and center point}
    \label{fig:brick_detection_model}
\end{figure}

Gripper and body cameras could in principle provide visual feedback for alignment verification. Computer vision-based object detection offers additional possibilities. Preliminary tests with an ATSS-MobileNetV2 model demonstrate brick identification in camera images (Figure~\ref{fig:brick_detection_model}). This suggests potential for adaptive component recognition and placement verification. This approach would generate a pick vector transformable from the camera frame to the gripper frame, potentially enabling grasping of components in unpredictable positions. Additionally, the computer vision model could be trained to identify construction tools to perform inventory management and tool tracking on construction sites.

Scaling to larger structures would require a multi-fiducial localization system or an alternative approach. Rather than increasing the effort required for placing fiducials, one option is to replace fiducial-based localization with visual \ac{SLAM}, though this introduces its own challenges in dynamic construction environments. Such a framework would form the basis for environmental modeling necessary for robust navigation. This would allow the robot to localize using a sparse map representation of the environment~\cite{spot_mapping}. This would reduce or eliminate the need for pre-recorded navigation, though autonomous path planning in active construction sites remains an open problem.

The system provides an initial step towards real-world masonry construction, with validation through clay bricks and mortar application as a direct extension. Further autonomous validation scenarios should include realistic construction environments featuring harsh terrain such as gravel and hard-to-reach areas. To support the \ac{MARA} workflow capabilities, diverse wall designs should be evaluated to test the robustness of the CAD-to-execution pipeline.

Multi-robot deployment could increase construction efficiency, supported by a complete \ac{ROS}~2 Humble-based system enabling collaborative map sharing for tasks such as site monitoring and inspection~\cite{gas_distribution_mapping}. Efficiency could also be improved by stowing multiple bricks on the robot's back, utilizing its remaining \SI{6}{\kilo\gram} payload capacity. 

The ability to traverse stairs may extend to movement on temporary scaffolding, which would decouple build height from the manipulator's reach. Possible further directions include real-time human-robot collaboration with live design updates, as-built verification against the target design model, and autonomous off-hours preparation of tools and materials.

The system reliably completed all experimental runs and demonstrated robust stair traversal between work zones. The overall performance highlights the feasibility of legged mobile manipulation for construction tasks, with occasional errors in the range of centimeters. These results suggest that such systems are a viable approach to automating on-site construction in unstructured, human-centric environments, where traditional robotic platforms have limitations.

\section*{Acknowledgement}
Funded by the German Research Foundation (DFG, Deutsche Forschungsgemeinschaft) as part of Germany’s Excellence Strategy – EXC 2050/2 – Project ID 390696704 – Cluster of Excellence “Centre for Tactile Internet with Human-in-the-Loop” (CeTI) of TUD Dresden University of Technology. The authors would like to thank the Federal Ministry of Research, Technology, and Space (BMFTR) for its support as part of the research program Communication Systems “Souverän. Digital. Vernetzt.”. Joint project 6G-life, project identification number: 16KIS2413K.

This paper was originally presented at ISARC 2026 and published in the Proceedings of the 43rd ISARC, 2026, Singapore, Singapore. DOI: \url{https://doi.org/10.22260/ISARC2026/0052}

\bibliography{ISARC}

\end{document}